# Dimensionality Reduction Ensembles

Colleen M. Farrelly
Independent Researcher
cfarrelly@med.miami.edu

**Abstract:**
Ensemble learning has had many successes in supervised learning, but it has been rare in unsupervised learning and dimensionality reduction. This study explores dimensionality reduction ensembles, using principal component analysis and manifold learning techniques to capture linear, nonlinear, local, and global features in the original dataset. Dimensionality reduction ensembles are tested first on simulation data and then on two real medical datasets using random forest classifiers; results suggest the efficacy of this approach, with accuracies approaching that of the full dataset. Limitations include computational cost of some algorithms with strong performance, which may be ameliorated through distributed computing and the development of more efficient versions of these algorithms.

**Introduction**

High-dimensional data is ubiquitous in data analysis today, and many of these datasets occupy a lower-dimensional space than that spanned by the full dataset (Van Der Maatan, Potsma, & Van den Herik, 2009). Many dimensionality reduction methods have been developed to identify this lower-dimensional space and map data to it, reducing the number of predictors in supervised learning problems and allowing for better visualization of data relations and clusters (Van Der Maatan, Potsma, & Van den Herik, 2009).

Principal component analysis (PCA) and manifold learning are two main approaches (Chatfield & Collins, 1980; Van Der Maatan, Potsma, & Van den Herik, 2009), the latter focusing on nonlinear subspace mapping rather than linear subspace mapping. Both approaches have had empirical success in correctly identifying subspaces that capture most of the data's variance, and many head-to-head comparisons have been made between these two approaches (Dupont et al., 2013; Plaza et al., 2005). One recent study showed the effectiveness of PCA on a variety of real datasets, suggesting that this is a good method overall (Van Der Maatan, Potsma, & Van den Herik, 2009).

However, the plethora of dimensionality reduction techniques provides a variety of nonlinear, linear, global, and local methods, and it is likely that each method captures different data features. Ensemble methods have achieved much success in supervised learning, from random forest to KNN regression ensembles to superlearners (Breiman, 2001; Farrelly, 2017). Ensembles exploit diversity and balance bias, variance, and covariance to achieve these results (Sollich & Krogh, 1996; Brown, Wyatt, & Tino, 2005), and it is likely that disparate dimensionality reduction methods will enhance diversity within a dimensionality reduction ensemble. Given these properties and successes, it is possible that creating an ensemble composed of local, global, linear, and nonlinear dimensionality reduction vectors will provide better joint embeddings than any single method, similar to how superlearner ensembles provide at least as good of prediction as any model component (Van der Laan, Polley, & Hubbard, 2007).

Very little has been explored with respect to ensembles in dimensionality reduction, and extant papers do not create ensembles with different base learner methods (Dupont & Ravet, 2013). For instance, the Dupont and Ravet attempt varies t-distributed Stochastic Neighbor Embedding (t-SNE) parameters in their ensemble, achieving better performance than a tuned t-SNE model, which, in turn, had achieved the best performance out of several dimensionality reduction techniques (Dupont et al., 2013). No attempts were made to create an ensemble using multiple techniques.

This paper aims to explore the efficacy of a mixed ensemble approach to dimensionality reduction on simulated data and two open-source medical datasets. Random forest models are used for classification on each ensemble, the full dataset, and each dimensionality reduction method's top two components, such that comparisons between approaches and the original data can be made. Medical datasets include the prediction of malignancy from the UCI Repository Breast Cancer Wisconsin (Original) Dataset (Mangasarian & Wolberg, 1990) and the prediction of cocaine, crack, and heroin usage from personality survey data (Fehrman et al., 2017). Results suggest that dimensionality reduction ensembles are an effective approach, outperforming the best individual dimensionality reduction approaches and achieving accuracy rivalling the results of the full dataset on the medical classification problems.

**Methods**

**I) Linear Approaches**

PCA is a linear dimensionality reduction technique based on projection methods, mapping a higher-dimensional Euclidean space to a lower-dimensional Euclidean space (Jolliffe, 1986). Given a data matrix, **X**, a target space, **Y**, and a projection matrix, **P**, PCA performs the following mapping:

$$Y = PX$$

The rows of **P** become a new basis for **X**, and by posing this equation as an optimization problem (maximizing variance, minimizing covariance between variables), this equation can be reformulated such that the singular value decomposition can be leveraged to solve the equation. The covariance matrix, **C**, can be expressed as:

$$C = \frac{1}{n-1} PXX^T P^T$$

Because the change of basis results in orthogonal bases and this method maximizes variance in the components, truncating the result will give a mapping to a lower-dimensional space using the new bases, while retaining as much variance from the original dataset as possible.

This method has relations to factor analysis and suffers from some of the same limitations (Farrelly et al., 2017). It will distort relationships and distances between points if either space is not linear, and results will not be as accurate as a method that treats those spaces as potentially nonlinear. Thus, methods incorporating geometric or topological properties may be more appropriate for some problems (Farrelly et al., 2017).

## II) Local Nonlinear Approaches

Local nonlinear methods focus on preserving neighborhood geometry when mapping points to lower-dimensional subspaces, and these subspaces may be curved or otherwise non-Euclidean. Many successes have come from these approaches, and some have guarantees about global properties, as well.

Locally-linear embedding (LLE) approaches the dimensionality reduction problem similarly to PCA but focuses on a neighborhood mapping, rather than a global mapping—including transformations, rotations, and rescaling dependent on geometrically-based weights (Roweis & Saul, 2000). Coordinates are chosen such that

$$Y = \sum_i \left(Y_i - \sum_j W_{ij} Y_j\right)^2$$

is minimized based on locally-linear errors based on the local weights on points i and j. The algorithm typically involves 3 steps: 1) choosing a point's k neighbors, 2) reconstructing these points using local weights, and 3) mapping to a new subspace. By knitting together neighborhoods, it is possible to infer some global geometric properties, and one advantage of this approach is its global optimization guarantees.

Hessian-based locally-linear embedding (HLLE) extends LLE through using manifold tangent space decompositions and least squares solvers (Donoho & Grimes, 2003). First, neighbors are identified and weighted as in LLE. From there, tangent coordinates are obtained through a singular value decomposition, and the Hessian is formed at each point through a least-squares solver. A symmetric matrix is built from these local neighborhoods, l, and their points, i and j, such that:

$$\boldsymbol{H}_{i,j} = \sum_l \sum_r \left(H^l_{r,i}, H^l_{r,j}\right)$$

From this, eigenanalysis of $\boldsymbol{H}_{i,j}$ can be performed to find the null space, which serves as an orthonormal basis. It has shown good empirical results on data known to be nonlinear, and obtains results similar to LLE and other manifold learning methods.

Laplacian eigenmaps (LE) is based on the connections between graph Laplacians, Laplace-Beltrami operators on a manifold, and the heat equation (Belkin & Niyogi, 2003). Because these are based on local properties of the manifold, the algorithm is fairly robust to noise, outliers, and global geometry. Like LLE and HLLE, this algorithm typically starts with a nearest neighbor graph, though graphs constructed through neighborhood size (distance between points) is possible. Weights are then chosen according to either connections between points or through the heat kernel defined between points $x_i$ and $x_j$:

$$W_{ij} = e^{-\frac{\|x_i - x_j\|^2}{t}}$$

Next, the algorithm computes eigenvalues and eigenvectors through the generalized eigenvector problem:

$$Ly = \lambda Dy$$

where **D** is the diagonal matrix of weights, **L** is the Laplacian matrix, and y corresponds to the solutions of this equation. This is similar to spectral clustering and graph-based methods, in general, which provide a wealth of theoretical results. In addition, many potential extensions exist with respect to simplicial complexes created from data (as opposed to graphs, which are 1-skeletons of higher-dimensional simplices) and higher-order Laplacians.

t-SNE is a local nonlinear dimensionality reduction method that implicitly incorporates multiple scales and maps these into a single representation (Van Der Maatan & Hinton, 2008). t-SNE extends stochastic neighbor embedding and ameliorates the issues related to the "crowding" phenomena, in which many points share a small space in the projection components. The algorithm starts by using conditional probability to represent similarities between points based on Gaussian distributions centered at a given point. Once a distance matrix is constructed, a low-dimensional representation is obtained through a gradient descent algorithm, which minimizes Kullback-Leibler divergence between all pairs of points. Kullback-Leibler divergence is inherently non-symmetric, and t-SNE imposes symmetry to improve results and utilizes a t-distribution rather than a Gaussian. Both t-SNE and SNE include a variance parameter for the distribution calculations, and this is related to the Shannon entropy of the probability distribution. Several papers have shown the efficacy of this approach to separate data relative to LLE and other manifold learning methods. A random-walk, random sample approach exists to reduce computational complexity, as does a tree-based representation (Van Der Maatan, 2014).

**III) Global Nonlinear Approaches**

Global nonlinear approaches aim to capture the full geometry of the data space in a mapping to lower-dimensional space, which may be non-Euclidean.

Kernel PCA (kPCA) extends the PCA algorithm to effectively capture nonlinear features or bases within the PCA framework (Weinberger, Sha, & Saul, 2004). The basic algorithms start with a mapping of the data to another pre-specified space, which is usually nonlinear. PCA is then performed on this new space (a kernel Hilbert space with nice, known geometric properties), and the results are taken as the new set of bases. Common mapping functions include Gaussian kernels, sigmoid kernels, radial basis function kernels, and linear kernels. Results are competitive with other manifold learning methods, and many algorithms, such as support vector machines, are related to this approach.

Multidimensional scaling (MDS) is a well-studied visualization technique that maps similarity scores between points (interpoint distances) to a lower-dimensional space for plotting, preserving those similarity scores as relative distances between points (Kruskal, 1964). When a Euclidean metric is used to measure similarity, this method corresponds with PCA exactly; it also has connections to Procrustes' analysis and correspondence analysis, making it ideal for data with categorical or ordinal predictors. When a different metric or a non-metric is used to create similarity scores, MDS extends PCA to a nonlinear mapping between spaces. Though originally used for visualization purposes, MDS provides a transformed set of bases that function as low-dimensional approximations of the data. However, MDS suffers from some of the same limitations of PCA, namely the inability to fully capture nonlinear features in the data.

Isometric feature mapping (ISOMAP) extends MDS to capture these nonlinear features, typically taking geodesic point distances as input (Tenenbaum, De Silva, & Langford, 2000). First, neighbors are connected either by connecting points within a certain distance or through a k-nearest neighbors design. Geodesic distances between points are obtained for all points in the neighbor graph. MDS is then applied to this matrix of distances to project them into a low-dimensional space. Coordinates are chosen that minimize

$$\|\tau(\boldsymbol{D}_G - \boldsymbol{D}_Y)\|_{L^2}$$

where $\boldsymbol{D}_G$ refers to the graph distance matrix, $\boldsymbol{D}_Y$ refers to the Euclidean space to which the data is mapped, $L^2$ refers to the matrix norm, and τ refers to a conversion between distances and inner products. ISOMAP asymptotically recovers the geometric structure of nonlinear manifolds, much like MDS and PCA do, and, as such, is an attractive option in dimensionality reduction. However, this algorithm is computationally extensive, and few alternative formulations exist in software packages.

## IV) Ensemble Design and Algorithm Parameters

Ensembles were created by running each dimensionality reduction algorithm in R on a simulated or real dataset, keeping the first two vectors of each algorithm's output, such that visualization plots would correspond well with the classification algorithm's input data. Each individual dimensionality reduction method's first and second components were used as a random forest model's input using the ranger R package default parameters, such that methods could be compared head-to-head in classification problems through an accuracy metric. Table 1 lists the algorithm parameters and R packages used for all considered dimensionality reduction methods.

*Table 1: Dimensionality reduction method details*

| Algorithm | R Package | Parameter Settings |
|---|---|---|
| PCA | base | Default |
| LLE | lle | k=12, reg=2, v=0.9, m=2 |
| ISOMAP | vegan | k=12, ndim=2 |
| kPCA | kernlab | kernel="rbfdot", kpar=0.2, features=2 |
| MDS | base | k=2 |
| HLLE | dimRed | knn=12 |
| t-SNE | dimRed | ndim=2, perplexity=80 |
| LE | dimRed | ndim=2 |

2 ensembles were created. The first utilized all dimensionality reduction output, such that the first 2 components of each algorithm were collected into a final dataset (large ensemble with 16 predictors), upon which the random forest model could learn. A smaller ensemble was also created, taking the first 2 components of MDS, t-SNE, and PCA results (small ensemble with 6 predictors); a random forest model was fit to this smaller ensemble for comparison with the original dataset, each individual dimensionality reduction methods, and the larger dimensionality reduction ensemble.

## V) Simulation Design

Simulated datasets included 13 variables (4 ordinal, 4 continuous, 5 binary), 4 of which had a predictive relationship with a binary outcome, with 2000 observations (such that all dimensionality reduction algorithms had converged). This reflects common industrial and medical datasets, where a variety of factors are collected—some of which are predictive and most of which are a mix of continuous and discrete measurements. Relationships simulated were 4 main effects relationships, 2 interaction effect relationships, and a mixed condition with 2 main effects relationships and 1 interaction effect relationship, yielding 3 types of datasets. 3 levels of noise were added to simulations, mimicking uncertainty and measurement error; Gaussian noise with a mean of 0 and variance of 0.25, 0.5, and 0.75, respectively, were added to the outcome for each dataset type. This gave a total of 9 simulated datasets. Simulations of each dataset were repeated 10 times, and accuracies for each model were averaged across conditions using a 70/30 train/test split.

**VI) Medical Classification Tasks**

Two medical datasets were used to compare dimensionality reduction techniques and to test the ensemble frameworks on real data. The first dataset is the UCI Repository Breast Cancer Wisconsin (Original) Dataset (https://archive.ics.uci.edu/ml/datasets/breast+cancer+wisconsin+(original)), which consists of an outcome (malignant/benign) and 9 predictors, some of which are nominal and some of which are continuous (clump thickness, uniformity of cell size, uniformity of cell shape, marginal adhesion, single epithelial cell size, bare nuclei, bland chromatin, normal nucleoli, and mitosis). This dataset includes 683 individuals with complete data, with 239 patients experiencing malignancy.

The second dataset, the Drug Consumption (quantified) Dataset (https://archive.ics.uci.edu/ml/datasets/Drug+consumption+%28quantified%29) includes 1885 individuals, and these individuals were classified as lifetime users/nonusers of 3 key illegal substances (cocaine, crack, and heroin) based on 12 continuous personality measures and categorical demographic factors (Big 5 traits, impulsivity score, sensation-seeking score, level of education, age, gender, country of current residence, and ethnicity). Lifetime cocaine use was reported in 45% of the sample (crack=14%, heroin=15%).

Analyses of these datasets followed the design detailed in Methods IV), such that each individual method's random forest model was compared to the two dimensionality reduction ensembles' random forest models and the original dataset's random forest model.

**Results**

**I) Simulations**

Across simulations, the full dataset random forest classifier attains the best accuracy, followed by the small ensemble classifier and the large ensemble classifier (Figure 1). MDS, PCA, and t-SNE generally perform well as individual methods but do not reach the accuracy of the full dataset or either of the ensembles; t-SNE appears to be the best method across most simulations, recapitulating previous results. HLLE performs exceptionally poorly on the simulation data, suggesting it either requires more dimensions to work well or does not perform well on this type of data.

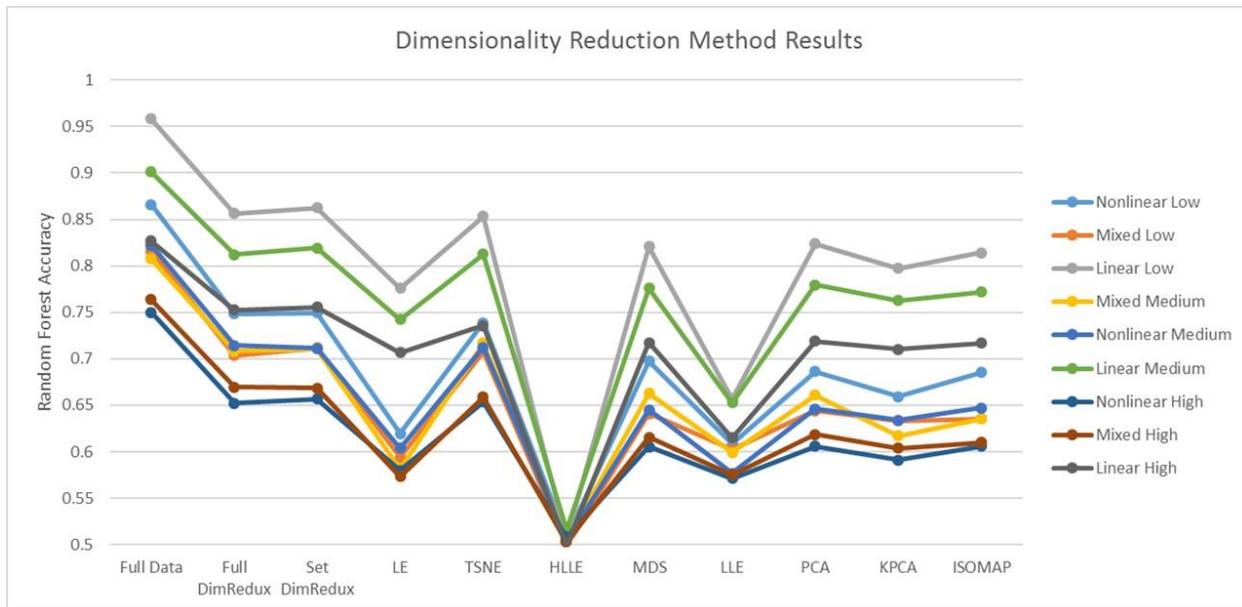

*Figure 1: Simulation results*

Separation is generally good across methods (Figure 2), though some methods are better than others.

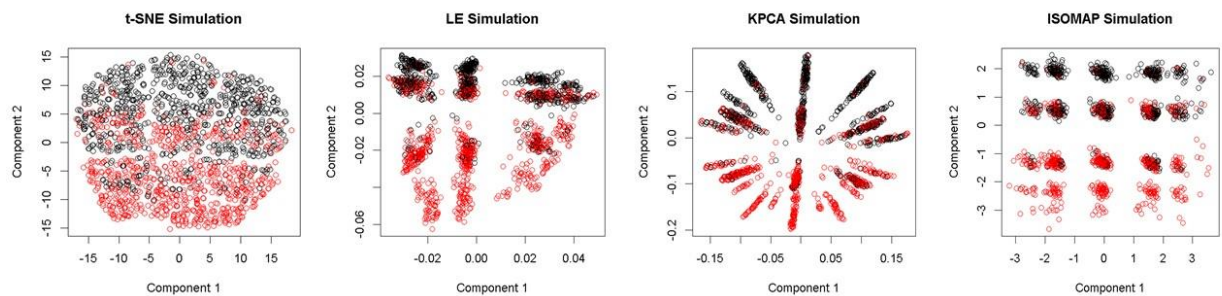

*Figure 2: Separation of simulation data*

## II) Medical Classification Datasets

On the Breast Cancer Wisconsin (Original) dataset, outcomes are well-separated using two components within most dimensionality reduction methods (Figure 3). In particular, t-SNE seems to separate data with very little overlap between groups. Again, t-SNE achieves the best performance of any single method, and it is outperformed by both ensemble methods (an additional 1-2% gain in accuracy), with the small ensemble slightly outperforming the large ensemble (Figure 4). On this dataset, both ensembles outperform the full dataset classifier, adding another 1% and 1.5% on accuracy. Again, HLLE struggles compared to other methods.

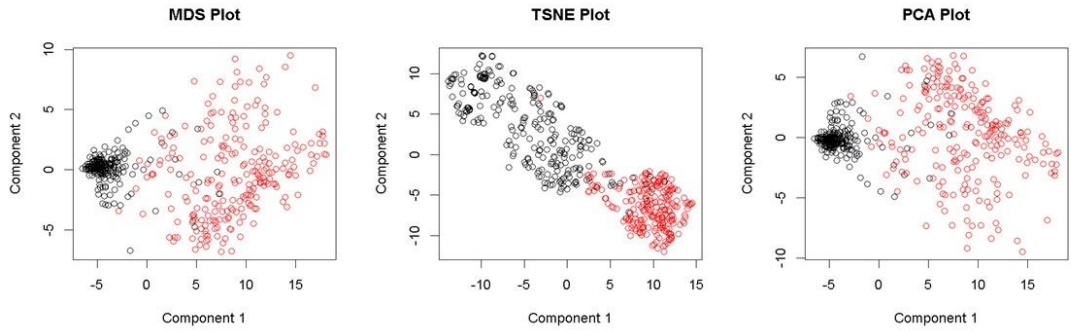

*Figure 3: Separation of Breast Cancer Wisconsin (Original) Dataset*

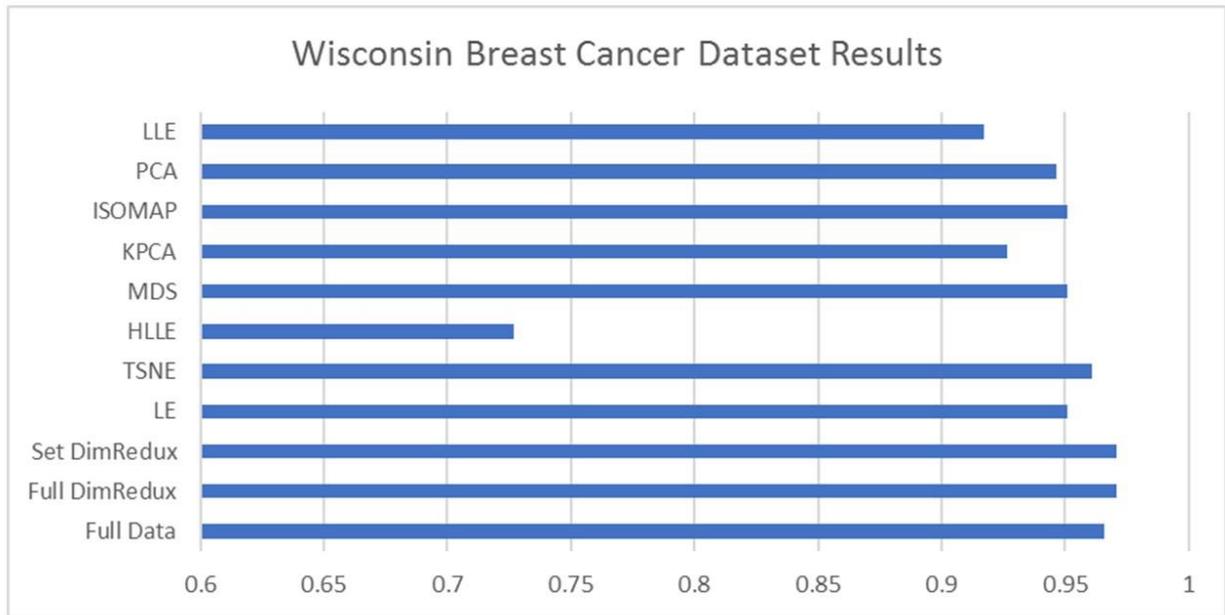

*Figure 4: Breast Cancer Wisconsin (Original) Dataset results*

The full dataset's results (Figure 5) suggest that the most important predictors of malignancy are uniformity of cell size, presence of bare nuclei, and uniformity of cell shape. Mitosis and marginal adhesion show almost no relationship with malignancy.

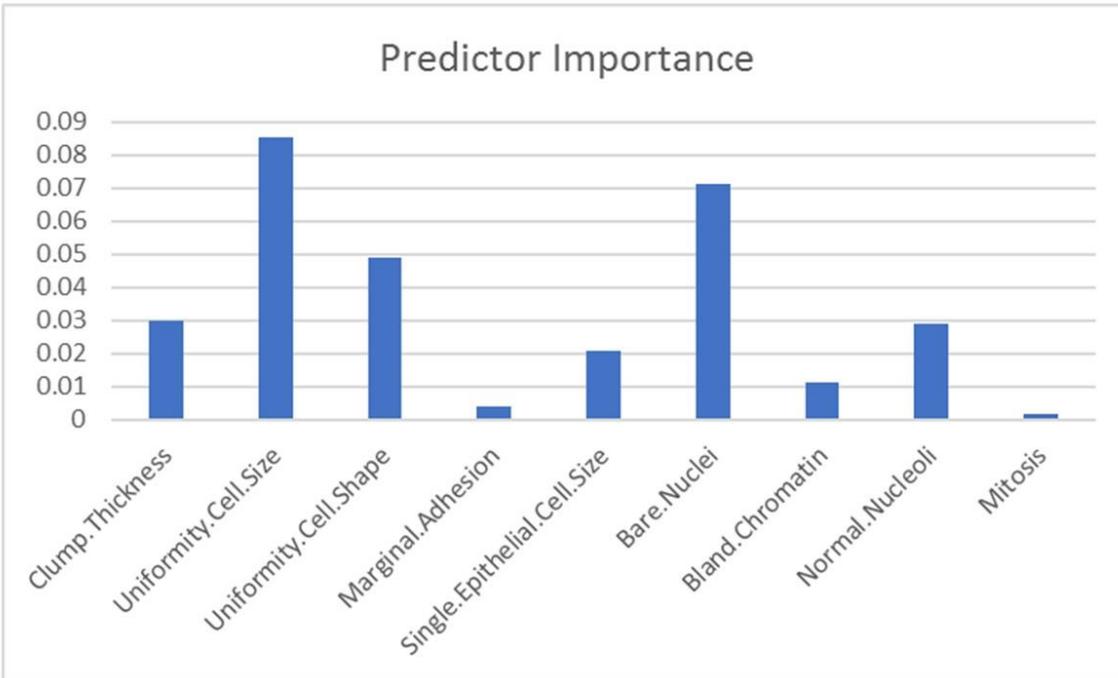

*Figure 5: Important predictors of malignancy*

On the three classification tasks involving the prediction of drug use, separation of outcomes on component plots is not well-defined for any method, suggesting that the data lives in a higher-dimensional space or that outcomes are not well-separated by this set of variables (Figure 6). With respect to prediction, the cocaine use model suggests large gains by ensembles, which fall a bit short of the full dataset's predictive model (Figure 7). Heroin and crack ensemble models achieve similar or improved predictive accuracy compared with the full dataset; the crack use models show substantial improvement over individual dimensionality reduction methods.

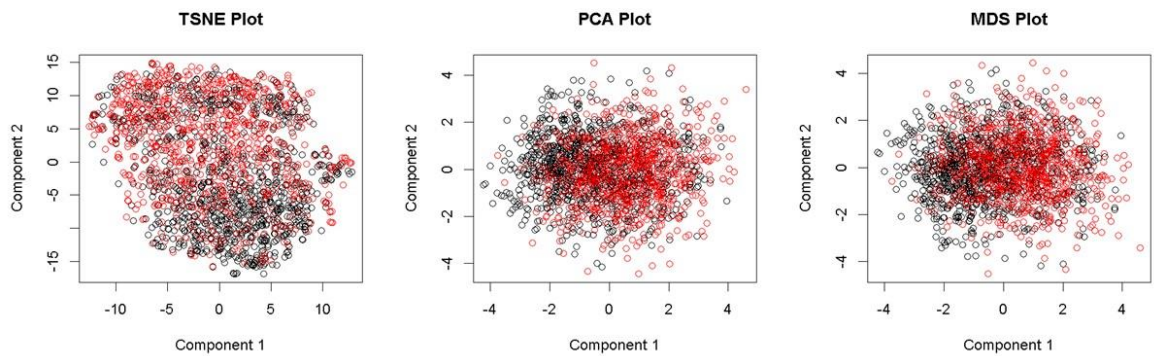

*Figure 6: Separation results for drug use dataset*

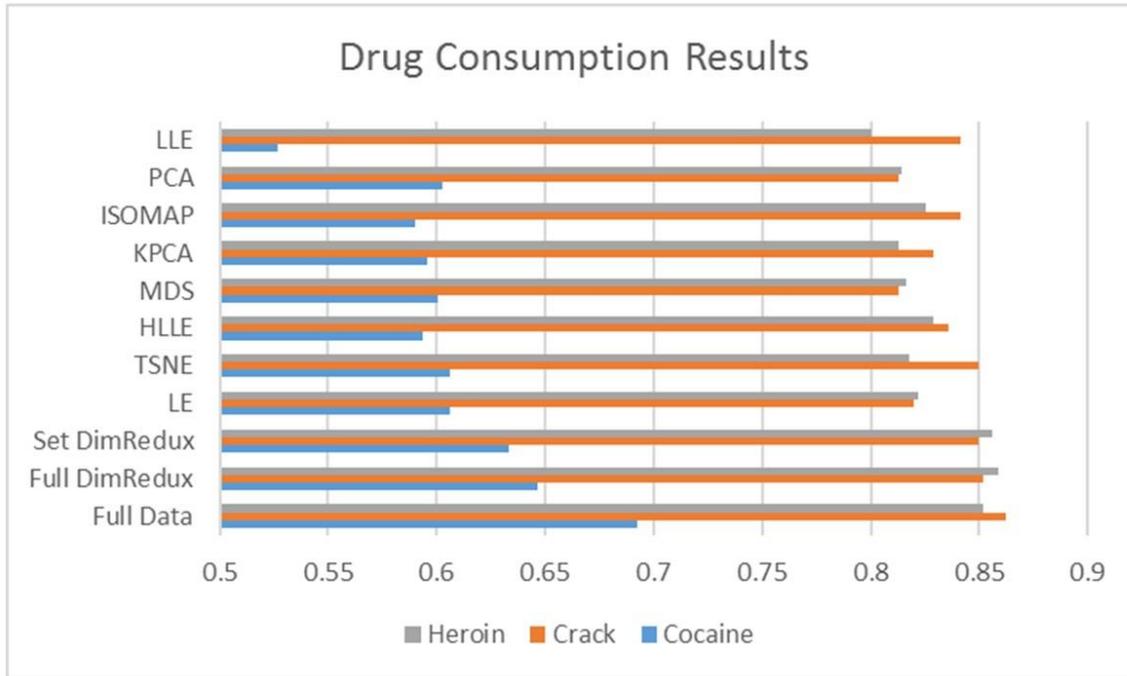

*Figure 7: Drug use dataset results*

The full drug use dataset's models suggest important predictors of drug use, including high openness and sensation seeking scores (particularly for cocaine use), low conscientiousness scores (for cocaine use only), and country in which one lives (Figure 8). Age also shows a relationship, with those reporting drug use generally younger than those who don't report drug use.

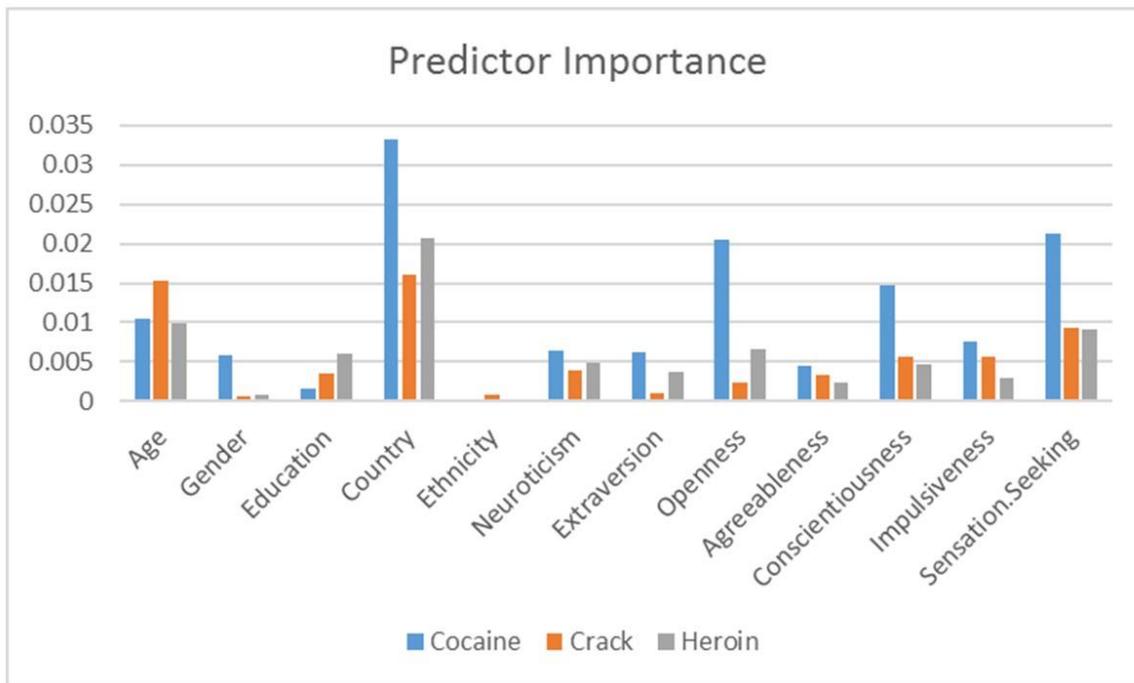

*Figure 8: Important predictors of drug use*

## Discussion

This study demonstrates the efficacy of dimensionality reduction ensembles in classification problems. Results from both simulations and real datasets suggest that these ensembles can outperform single dimensionality reduction methods and can approach or surpass classification accuracy using the initial dataset's predictors. This suggests that a diverse ensemble of dimensionality reduction components can be useful on complex datasets or high-dimensional datasets to reduce computational load on predictive models, to visualize subpopulations, and to improve predictive performance.

One limitation of many manifold learning methods is their computational complexity. This limits their applicability to large datasets and is a major drawback when creating ensembles, particularly ensembles using t-SNE or ISOMAP (Van Der Maatan, 2014; Tenenbaum, De Silva, & Langford, 2000). Sometimes, the most accurate local or global method is too computationally expensive for practical use. Efforts to reduce computational cost have been successful for some algorithms (particularly t-SNE) and are on-going (Van Der Maatan, 2014), but algorithms may still be infeasible for some datasets. However, given the nature of ensemble construction, diverse ensembles can still be created by choosing more computationally-feasible approaches to local or global manifold learning.

Another limitation is convergence; this study bypassed the issue by using large enough sample sizes to induce convergence. For small sample sizes, some algorithms may not have sufficient examples upon which to learn the manifold, limiting their use on small samples; other algorithms, such as the tools of topological data analysis, may be more appropriate than dimensionality reduction or machine learning methods in some of these cases (Farrelly et al, 2017).

However, these results point towards an effective approach to dimensionality reduction and data visualization. As new methods and new approaches to dimensionality reduction are developed, diversity in these ensembles can increase, leading to better prediction and visualization.